\documentclass[lettersize,journal]{IEEEtran}
\usepackage{amsmath,amsfonts}
\usepackage{algorithmic}
\usepackage{algorithm}

\usepackage{array}
\usepackage{textcomp}
\usepackage{stfloats}
\usepackage{url}
\usepackage{verbatim}
\usepackage{graphicx}
\usepackage{cite}
\usepackage{multirow}
\usepackage{subfigure}
\usepackage{gensymb}
\usepackage{amssymb}
\usepackage{color}
\usepackage{hyperref}

\usepackage{flushend}

\usepackage[capitalize]{cleveref}
\crefname{section}{Sec.}{Secs.}
\Crefname{section}{Section}{Sections}
\Crefname{table}{Table}{Tables}
\crefname{table}{Tab.}{Tabs.}

\begin{document}

\title{ReSmooth: Detecting and Utilizing OOD Samples when Training with Data Augmentation}

\author{Chenyang Wang,
        Junjun Jiang,~\IEEEmembership{Senior Member,~IEEE,}
        Xiong Zhou,
        Xianming Liu,~\IEEEmembership{Member,~IEEE}
        % <-this % stops a space
\thanks{C. Wang, J. Jiang, X. Zhou and X. Liu are with the School of Computer Science and Technology, Harbin Institute of Technology, Harbin 150001, China. E-mail: \{cswcy, jiangjunjun, cszx, csxm\}@hit.edu.cn.}% <-this % stops a space
%\thanks{Manuscript received April 19, 2021; revised August 16, 2021.}
}

% The paper headers
\markboth{Journal of \LaTeX\ Class Files,~Vol.~14, No.~8, August~2021}%
{Shell \MakeLowercase{\textit{et al.}}:ReSmooth: Detecting and Utilizing OOD Samples when Training with Data Augmentation}

\IEEEpubid{0000--0000/00\$00.00~\copyright~2021 IEEE}
% Remember, if you use this you must call \IEEEpubidadjcol in the second
% column for its text to clear the IEEEpubid mark.

\maketitle

\begin{abstract}
Data augmentation (DA) is a widely used technique for enhancing the training of deep neural networks. Recent DA techniques which achieve state-of-the-art performance always meet the need for diversity in augmented training samples. However, an augmentation strategy that has a high diversity usually introduces out-of-distribution (OOD) augmented samples and these samples consequently impair the performance. To alleviate this issue, we propose ReSmooth, a framework that firstly detects OOD samples in augmented samples and then leverages them. To be specific, we first use a Gaussian mixture model to fit the loss distribution of both the original and augmented samples and accordingly split these samples into in-distribution (ID) samples and OOD samples. Then we start a new training where ID and OOD samples are incorporated with different smooth labels. By treating ID samples and OOD samples unequally, we can make better use of the diverse augmented data. Further, we incorporate our ReSmooth framework with negative data augmentation strategies. By properly handling their intentionally created OOD samples, the classification performance of negative data augmentations is largely ameliorated. Experiments on several classification benchmarks show that ReSmooth can be easily extended to existing augmentation strategies (such as RandAugment, rotate, and jigsaw) and improve on them. Our code is available at \href{https://github.com/Chenyang4/ReSmooth}{https://github.com/Chenyang4/ReSmooth}.
\end{abstract}


\section{Introduction}
\IEEEPARstart{D}{ata} augmentation is an essential and effective technique for the learning problem. By virtue of it, many deep learning methods can substantially boost their performance in various vision tasks, including object recognition\cite{devries2017improved, cubuk2018autoaugment, cubuk2020randaugment}, defect detection\cite{9732170}, clustering\cite{ji2019invariant, van2020scan, 9410431}, object detection\cite{girshick2018detectron} and semantic segmentation\cite{fang2019instaboost}. There are two main intuitions in the literature to design a good DA strategy, namely \emph{minimal data distribution shift} and \emph{maximal diversity of augmented samples}. Typical examples for the former include image transformations like rotation, flipping, and cropping\cite{krizhevsky2012imagenet}, while the latter mainly includes methods considering the combination of multiple transforms~\cite{cubuk2018autoaugment,cubuk2020randaugment,ratner2017learning}. According to the empirical results in \cite{gontijo2020affinity}, a better augmentation policy should have both bigger affinity (between the augmented data distribution and original data distribution) and bigger diversity (of augmented samples).

However, recently proposed DA strategies that achieve state-of-the-art performance usually ignore the former and mainly focus on the diversity property, which we call diverse DA. As a result, some samples are inevitably out of the original data distribution, leading to a decline in affinity. For these samples, one feasible solution is to discard them during training, but this will attenuate the diversity of the training samples and consequently limit the model performance (see \Cref{sec:need} for details). Therefore, we wonder if there is a strategy to maintain the diversity of the augmented samples while not detrimentally affecting the learning of samples lying in original data distribution.
\IEEEpubidadjcol

\begin{figure*}
  \centering
  \includegraphics[width=1.0\linewidth]{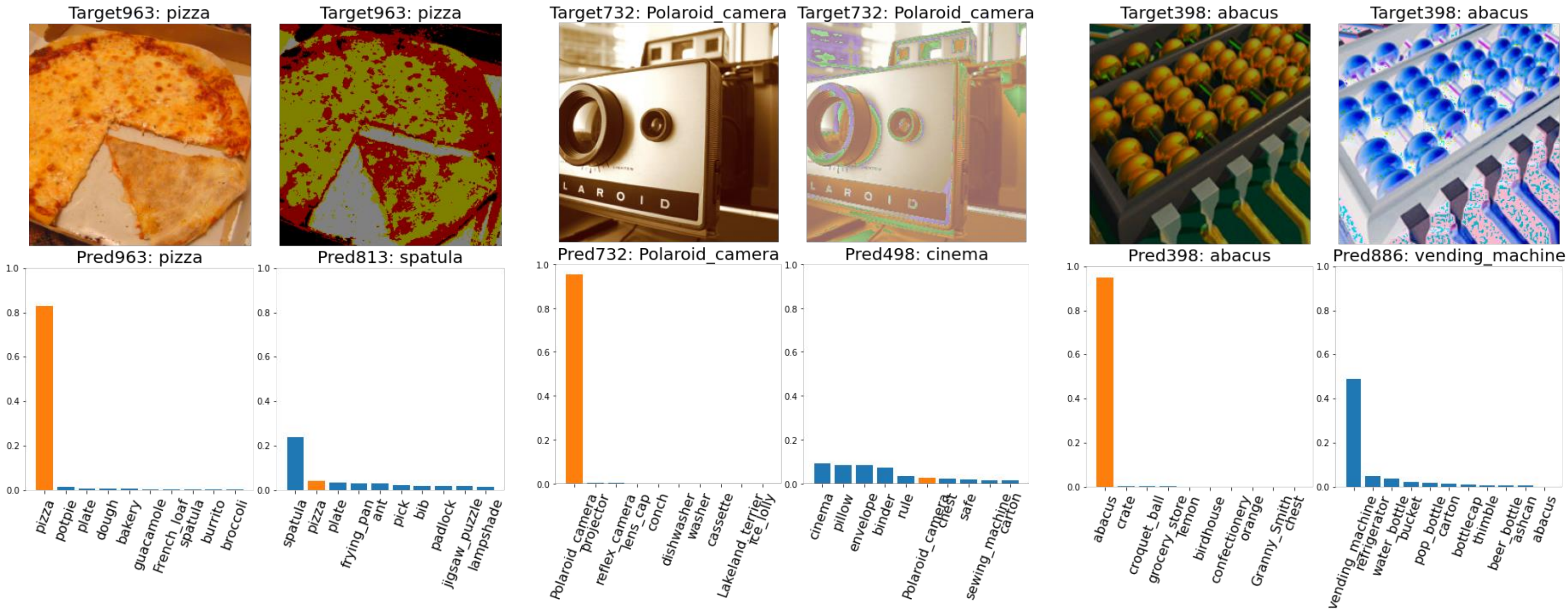}
  \caption{Some examples of DAOOD samples in ImageNet training set. For each sample, we show the image before and after augmentation in the upper left and upper right, the lower left and right are their predictions by the model pretrained on unaugmented data.}
  \label{fig:DAOOD}
\end{figure*}

To that effect, we dig into these OOD samples and notice an interesting phenomenon: a big part of these OOD samples, unlike the announcements in previous works~\cite{wei2020circumventing,gong2021keepaugment}, are not noisy or ambiguous but semantic-unchanging for the target task, which means these samples do not change their task-related attributes (e.g., labels in the context of the classification task) from the view of us. We call these samples DA-incurred OOD (DAOOD) samples and some examples on ImageNet are shown in \Cref{fig:DAOOD}. This phenomenon widely exists in natural image datasets, but treating DAOOD samples and ID samples (including both the original samples and data augmented ID (DAID) samples) naively equal will impair the fit of ID samples and thus result in sub-optimal performance. In this paper, we mainly focus on leveraging these samples with our proposed ReSmooth framework.

ReSmooth adopts a detecting-utilizing two-step scheme to make use of DAOOD samples. In the first step, we try to divide the training samples into ID samples and DAOOD samples. In the second step, we differently treat the two groups of the samples where, ideally, the introduction of DAOOD samples will benefit the learning of ID samples. To be specific, we first train a network with unaugmented samples and then use the trained network to estimate a mixture model by loss distribution modeling. Then, the mixture model is used to distinguish OOD samples from ID samples. With this separation, ID samples are trained by the original training scheme with standard cross-entropy loss, whereas OOD samples are trained with label smoothing. The per-sample smoothness parameter is acquired according to the posterior probability predicted by the estimated mixture model. With this divide-and-conquer strategy, our method can make full use of both the original data and the newly obtained augmented data.

Unlike diverse augmentation, some known transformations in the literature are rarely used in regular training pipelines for the reason of poor performance or unsatisfying interpretability, e.g., rotation with large angle as in \cite{gidaris2018unsupervised} and jigsaw transformation as in \cite{noroozi2016unsupervised}. These transformations are also known as negative DA (NDA)~\cite{sinha2021negative} which pursues intentionally creating out-of-distribution samples. Though not lying on the support of the data distribution, samples created in this way are informative and can be semantic-unchanging for the target task. They can be viewed as DAOOD samples from another source and be leveraged by the proposed ReSmooth.

The main contributions of this work are summarized as follows:
\begin{itemize}
\item We reveal an interesting phenomenon that lots of OOD samples caused by diverse DAs are semantic-unchanging and can be exploited for training.
\item We propose a novel and simple framework for learning with diverse and negative DA. The framework can be flexibly adopted by combining different DA strategies, data split methods, and OOD utilizing designs.
\item By combining with different DA strategies, we empirically show that our method can further improve on the existing SOTA DA strategies. 
\end{itemize}

The rest of the paper is unfolded as follows. \Cref{sec:Related} introduces previous works related to ReSmooth, including related data augmentation strategies, data split methods and learning strategies. \Cref{sec:Method} introduces the adopted data augmentation techniques in our work, gives the definition of DAOOD samples and presents the proposed framework, and \Cref{sec:Experiments} shows the comparison results of our proposed method with others, ablation analysis and some potential concerns. \Cref{sec:Conclusions} concludes this study. (\Cref{tab:abbreviations} is provided for convenience of reading the rest of the paper.)

\section{Related Work}

\begin{table}[t]
\centering
\caption{Abbreviations in this paper for better reading.}
\begin{tabular}{cc}
\hline
Abbreviation  & Full name                              \\ \hline
DA            & data augmentation                      \\
NDA           & negative data augmentation             \\
OOD samples   & out-of-distribution samples            \\
ID samples    & in-distribution samples                \\
DAOOD samples & DA-incured OOD samples                 \\
DAID samples  & in-distribution data augmented samples \\
LS            & label smoothing                        \\
LSR           & label smoothing regularization         \\
SA            & standard augmentation                  \\
RA            & RandAugment                            \\
AA            & AutoAugment                            \\
GMM           & Gaussian Mixture Model                 \\ \hline
\end{tabular}
\label{tab:abbreviations}
\end{table}

\label{sec:Related}
Our work aims at detecting and utilizing OOD samples when training with data augmentation. In other word, the key contribution of our work is the data separating strategy and the following learning strategy under the background of data augmentation. In this section, we retrospect the  related works of data augmentation, data separating methods and learning strategies with separated data.

\subsection{Data Augmentation}
In this paper, we mainly focus on the DA which pursues the diversity property. AutoAugment\cite{cubuk2018autoaugment} is a typical diverse data augmentation method for searching the optimal combination of basic transformations by reinforcement learning. Following the idea of auto augmenting, some efforts are made to alleviate the prohibitive computation cost of AutoAugment. To name some of many, population-based augmentation\cite{ho2019population} applied evolution algorithm to generate the augmentation policy, RandAugment\cite{cubuk2020randaugment} utilized a reduced search space for cost-free search, and AutoDo\cite{gudovskiy2021autodo} modeled the parameters in augmentation as a source of hyperparameters and optimize them with scalable probabilistic implicit differentiation. However, these augmentation strategies may introduce low-quality samples and harm the performance. To that effect, AugMix~\cite{hendrycks2019augmix} tried to output natural-looking samples, \cite{gong2021keepaugment} proposed to keep the fidelity of the augmented samples by pasting the patches back from the original samples or avoiding cutting the important areas, and \cite{suzuki2022teachaugment} introduced teacher knowledge to constrain adversarial data augmentation strategies. What is more, Fast AutoAugment~\cite{lim2019fast} and Faster AutoAugment~\cite{hataya2020faster} are two follow-up researches of AutoAugment which explicitly match the density between augmented and unaugmented data. 

The work most closely related to ours is \cite{wei2020circumventing}. In \cite{wei2020circumventing}, like ReSmooth, a network is firstly trained with unaugmented data and then the augmented data are inputted to the network to get the predictions. The key differences lie in how to train a new better network given the predictions. \cite{wei2020circumventing} builds upon the observations that when heavy data augmentation is added to the training image, it is probable that part of its semantic information is removed. As a result, insisting on the ground-truth label is no longer the best choice and the predictions may represent the true semantics left in the data. Therefore, \cite{wei2020circumventing} propose to distill the new network with the predictions. In ReSmooth, however, we just use the predictions to estimate a loss distribution and softly split the data accordingly for a new training, because we argue that DAOOD samples are the most of cases in augmented data and distillation will harm the learning of them. We provide more analysis of the augmented data in \Cref{sec:ood} and give the possible reason why we perform better than \cite{wei2020circumventing} there.

\subsection{Heterogeneous Data Separating}
Real-world data is collected from multiple sources and therefore is always heterogeneous. A branch of methods handling this situation is to separate the heterogeneous data into homogeneous splits based on some statistics during training or from extra prior knowledge. For the conventional OOD detection setting~\cite{yang2021generalized}, a pretrained model is provided by solving tasks on the ID samples and a score is needed for the detection split. Based on the pretrained model, \cite{hendrycks2016baseline} used maximum Softmax score, \cite{lee2018simple} adopted the maximum Mahalanobis distance of the penultimate layer representations, and \cite{liu2020energy} proposed the energy score to measure the sample energy. For the task of noisy label learning, \cite{li2020dividemix,arazo2019unsupervised} proposed to model the data noise according to the loss statistics during training. They suggest that the loss distributions are separable during the early stage of training. Different from them, \cite{wei2020combating, han2018co, malach2017decoupling} proposed to identify the samples with noisy labels directly from the loss values of two separated models. In addition to loss statistics, some methods explores to use more complicated and effective statistics, like Area under the Margin Ranking\cite{pleiss2020identifying}. In unsupervised representation learning task with imbalanced data, \cite{hooker2019compressed, jiang2021self} identified hard-to-memory samples from tail classes by different outputs between the network and its pruned version. Some works used a threshold of the loss to distinguish atypical data\cite{wen2021seeking} or not well-learned data\cite{sohn2020fixmatch} in voice-face mapping and semi-supervised learning, respectively. Here, we mainly focus on the heterogeneous data created by DA and follow the idea of noise distribution modeling to separate the ID and DAOOD samples.

\subsection{Learning Strategy with Separated Data}
Different tasks treat separated samples differently. For example, in noisy label learning task, identified noisy samples can be dropped or treated as unlabeled samples. In the field of DA, augmented data can be viewed as naturally separated data compared to original data. Based on this assumption, \cite{merchant2020does} proposed to use split BatchNorms to capture different statistics. \cite{yi2021reweighting} used a much more fine-grained strategy considering per-sample per-augmentation reweighting with a min-max optimization scheme. Different from these methods, in this work, we consider the relationships among samples augmented by the same strategy. We follow the assumption that the property of an augmented sample is a function of both the sample and the augmentation strategy. So even given a specific DA, the learning strategy should vary across different samples, especially for diverse augmentation strategies like RandAugment \cite{cubuk2020randaugment}.

\section{Method}
\label{sec:Method}
In this section, we first introduce DA techniques considered in this work including diverse DA and negative DA. Then we give the definition of so-called DAOOD samples. Finally, we introduce our ReSmooth framework and specific detection and utilizing strategies.

\subsection{Preliminaries: Data Augmentation}
\begin{figure}[t]
  \centering
  \includegraphics[width=1.0\linewidth]{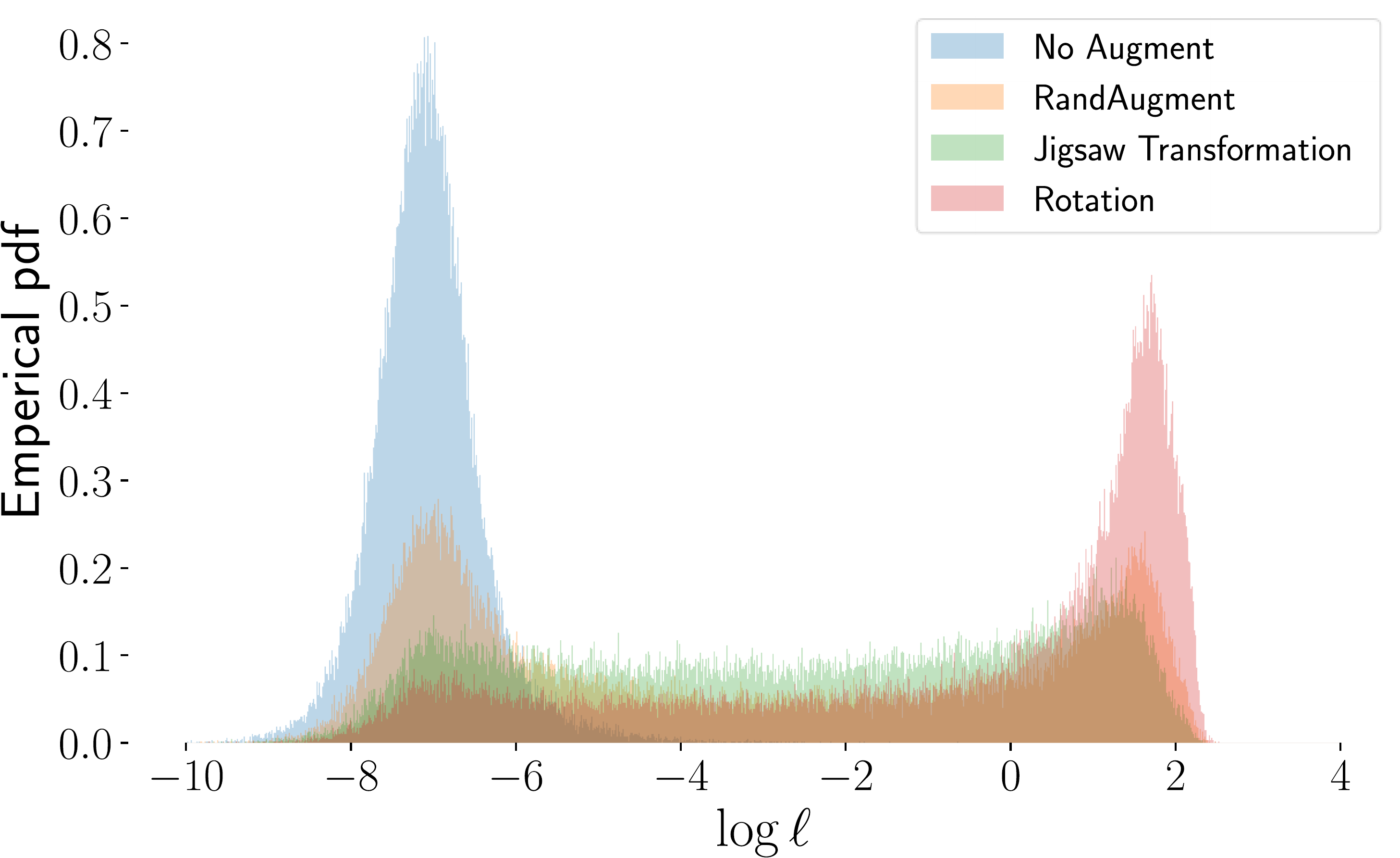}
  \caption{Loss distributions of different augmentation strategies on CIFAR-10 training set. The predictions for calculating losses are from the model trained on the original unaugmented data.}
  \label{fig:nda}
\end{figure}

We focus on two main kinds of DA: diverse DA and negative DA. Diverse DA implies the augmentation that creates a much more diverse training set throughout the training phase, while negative DA refers to the augmentation that aims to create OOD samples during training. In this work, we propose to estimate data distributions in the loss space (see \Cref{sec:daood}) and thus diverse DA and NDA can be further depicted with the help of loss statistics. In this sense, diverse DA refers to the augmentation where augmented samples are of a big variance/diversity in the loss space. Specifically, a part of augmented samples of the diverse DA lie in the original data distribution while another part of augmented samples are not. For augmented samples of the NDA, the situation changes, that most of augmented samples are heavy out-of-distribution and lie far away from the original data distribution. To be noticed, both diverse DA and NDA considered in this work are ideally semantic-unchanging for the classification task.

We choose RandAugment (RA) \cite{cubuk2020randaugment} as the representative of diverse DA and choose jigsaw and rotation as the representatives of NDA in our experiments. Loss statistics of them can be found in \Cref{fig:nda}.\\
{\bf RandAugment} considered the combination of multiple known transformations for image data, such as contrast, sharpness and solarize. To make the transformed images meaningful for data augmentation, RA designed a largely reduced search space to find the optimal parameters to combine the transformations. Because of its probabilistic augmentation property, the augmented samples are of a big diversity.\\
{\bf {\em Jigsaw} transformation} is a transformation that rearranges the spatial order of patches. It is a representative transformation that destroys the spatial correlation of images but preserves the local information. Though it is out of distribution, viewing it as the positive DA can help the model to distinguish object parts and/or resemble parts to recognize it.\\
{\bf Rotation transformation} is also semantic-unchanging except for the rotation angle. It can be viewed as a kind of NDA for the reason that samples after rotation with big angles ($90\degree, 180\degree, 270\degree$) are heavy out-of-distribution. It may relate to the inductive bias of the ConvNet which doesn't have the rotation-invariant property. Whatever, rotated images follow the definition of DAOOD and can be utilized to boost the performance on the original dataset.

\subsection{Data Augmentation-incurred OOD}
\label{sec:daood}

We begin with the formulation of the learning problem with DA. Let $\mathcal{X}$ denote the data space and $\mathcal{Y}=\{1,\cdots,K\}$ denote the label set. We are supposed to learn a labeling function $f: \mathcal X \xrightarrow{} \mathcal Y$ on the dataset $\mathcal D=\{(x_i, y_i)\}_{i=1}^N$ drawn from a distribution $\mathcal P$ over $\mathcal{X}\times\mathcal{Y}$. Given a loss function $\ell$, we usually minimize the empirical risk of $\ell$ over the observed dataset $\mathcal D$:

\begin{equation}
\label{eq:1}
R_{emp}\left(f\right) = \frac{1}{|\mathcal D|}\sum_{(x_i, y_i) \in \mathcal D} \ell\left(f\left(x_i\right), y_i\right).
\end{equation}

Let $\mathcal{A}(\cdot|x)$ denote the distribution of the transformations in a data augmentation strategy and $\mathcal D'$ denote the augmented dataset. The sample in $\mathcal D'=\{(x_i', y_i')\}_{i=1}^{N'}$ follows: ${(x_i', y_i') = (t(x_i), y_i)}, t\sim \mathcal{A}(\cdot|x_i)$. Though minimizing \Cref{eq:1} on $\mathcal{D'}$ is tenable and most of the existing methods are based on the objective, it is hard to optimize in practice. We find the main reason behind this phenomenon is the introduction of DAOOD samples. Here, we refer to DAOOD samples as the augmented samples with semantics unchanged but model-related metric changed. In other words, DAOOD samples are augmented samples where semantic information, especially task-related information, is unchanged from the view of human-being but changed from the view of the model.

To further illustrate that point, we compare it with noisy label learning. Data with noisy labels always introduce harmful knowledge for learning but data augmented mainly introduces new knowledge for better generalization. As a result, learning with DAOOD is not like learning with noisy data that harm performance. The key lies in how to generalize to the introduced knowledge without confusing the model or impairing the learning of existing knowledge in the original dataset.

\subsection{ReSmooth: DAOOD Detection and Utilization}
\label{sec:du}
\begin{figure}[t]
  \centering
  \includegraphics[width=0.9\linewidth]{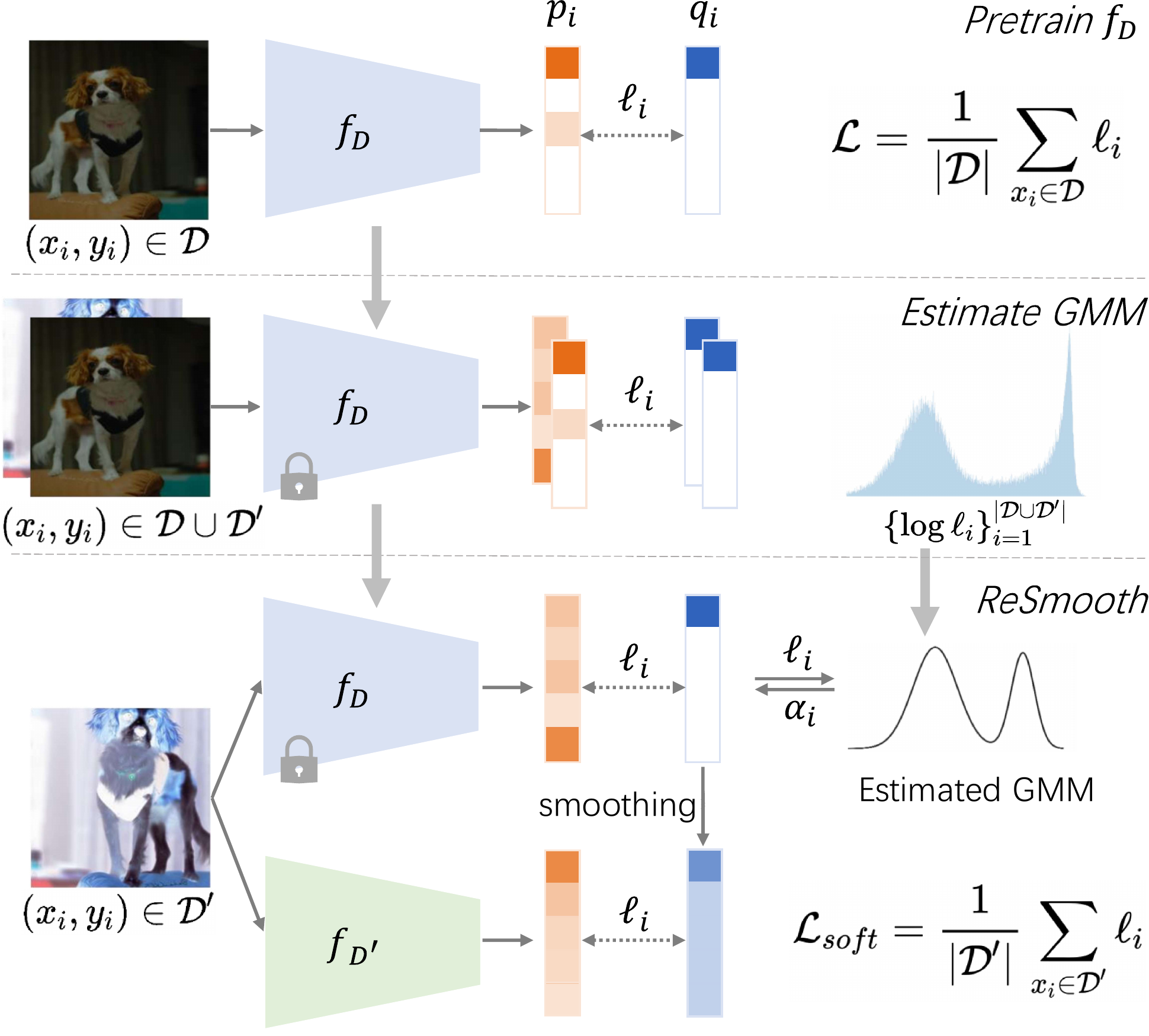}
  \caption{An overview of our proposed ReSmooth framework. Firstly, we train a model $f_D$ on the unaugmented dataset to learn the statistics of the original data distribution implicitly. By it, we can draw the mixture of the original distribution and OOD distribution in the $\log\ell$ space and estimate a GMM model to fit it. Then in the ReSmooth training stage, we can predict posterior probability $w_i$ of each input and define the smooth parameter linearly by $\alpha_i=\alpha\cdot w_i$. Using these sample-wise smooth parameters, we can train the network according to \Cref{eq:8}.}
  \label{fig:framework}
  \vspace{-0.4cm}
\end{figure}

Our proposed ReSmooth framework firstly detects DAOOD samples by leveraging a model pretrained on unaugmented data and an estimated GMM in the loss space of the pretrained model. Then we utilize the detected OOD samples to train a new model by sample-wise label smoothing. An overview of our framework is in \Cref{fig:framework}.

\subsubsection{DAOOD Detection}

The first part of ReSmooth framework is to identify DAOOD samples from the augmented training set. Formally, we assume DAID samples are lying in the original data distribution $\mathcal P$ while DAOOD samples are not. In our framework, we choose to approximately estimate $\mathcal{P}$ in the loss space because DAOOD samples are samples recognized wrongly by a model learned on $\mathcal P$ and thus have large loss values. Notably, a trained model is often sensitive to the distribution of the training data~\cite{gontijo2020affinity}. Therefore, we first train a model $f_{\mathcal D}$ on unaugmented dataset $\mathcal D$ and then use $f_{\mathcal D}$ to acquire the loss distribution of $\mathcal D\cup \mathcal D'$. 

\begin{figure*}[t]
  \centering
  \includegraphics[width=1.0\linewidth]{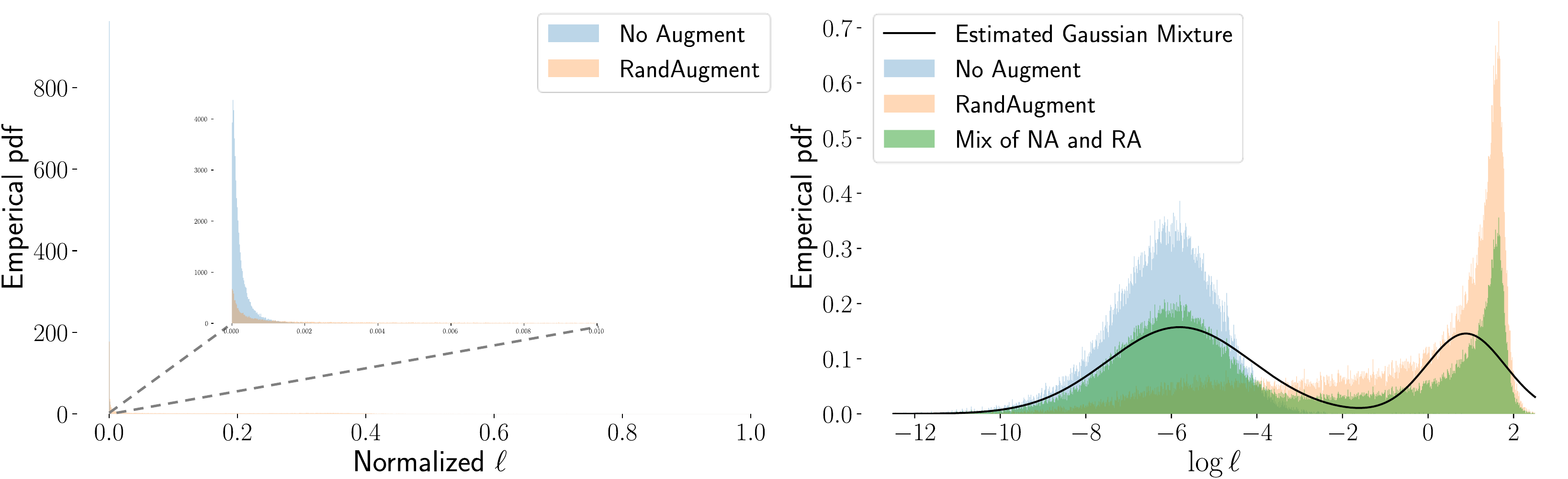}
  \caption{Left: Normalized loss distribution on CIFAR-100 training set and the zoomed-in axis is provided for better view. Right: Log loss distribution on CIFAR-100 training set. By comparison, it can be found that the log loss can better distinguish the two distributions and can be modeled by the Gaussian Mixture Model.}
  \label{fig:GMM}
\end{figure*}

At the first glance, it is hard to distinguish what $\mathcal P$ is like for the reason that the loss distribution exhibits high skew to zero and is of a long tail. When we view loss values in the logarithm domain, however, the loss distribution is mostly like a Gaussian mixture distribution~\cite{permuter2006study} (as shown in \Cref{fig:GMM}):
\begin{equation}
\label{eq:2}
\log\mathcal{L} \sim \pi_0 \mathcal N(\mu_0, \sigma_0) + \pi_1 \mathcal N(\mu_1, \sigma_1).
\end{equation}
where $\mu_i$, $\sigma_i$ is the mean and variance of the $i$th Gaussian component $\mathcal N(\mu_i, \sigma_i)$ and $\pi_i$ is its mixing coefficient, satisfying $\sum_{i}{\pi_i}=1$. Notably, $\mathcal N(\mu_0, \sigma_0)$ corresponds to the original data distribution $\mathcal P$ in the loss space, which satisfies $\mu_0=\min_i\{\mu_i\}$. We estimate $\boldsymbol{\mu, \sigma,\pi}$ by the EM algorithm. Then, we can estimate the posterior probability $w_i$ of a sample $x_i$ belonging to $\mathcal N(\mu_0, \sigma_0)$ given its loss value $\ell_i$:
\begin{equation}
\label{eq:3}
w_i\triangleq p(x_i\in {\mathcal P}|\ell_i)=\frac{\pi_0 p_{\mathcal N(\mu_{0}, \sigma_{0})}(\log \ell_i)}
{\sum_{k=0}^1{\pi_k p_{\mathcal N(\mu_k, \sigma_k)}(\log \ell_i)}}.
\end{equation}

We finish the detection stage by getting the posterior probabilities, where we can softly split samples (or explicitly split samples into $\mathcal D'_{ID}$ and $\mathcal D'_{OOD}$ by setting a hard threshold $\tau$). Note that the detection procedure is in an online way because the DA is applied randomly during training and even the effect of the same augmentation function differs for different samples.

\subsubsection{DAOOD Utilization}

\begin{algorithm}[t]
\caption{ReSmooth}
\label{alg:resmooth}
\begin{algorithmic}[1]
\REQUIRE $\alpha$, $\mathcal{D}$, $\mathcal{D'}$
\ENSURE $f_\mathcal{D'}$
\STATE Randomly initialize $f_\mathcal{D}$
\FOR {epoch=1,2,$\cdots$,N}
\FOR {j=1,2,$\cdots$,k} 
\STATE Sample a mini-batch data $X_j$ from $\mathcal{D}$  
\STATE Compute loss $\mathcal{L}$ on $X_j$ with $f_\mathcal{D}$ as in \cref{eq:1}
\STATE Update $f_\mathcal{D}$ with $\mathcal{L}$
\ENDFOR
\ENDFOR \hfill // Pretraining $f_\mathcal{D}$
\STATE $S=\varnothing$
\FOR {$(x_i, y_i)\in\mathcal{D} \cup \mathcal{D'}$}
\STATE $p_i=f_D(x_i), q_i=one\-hot(y_i)$
\STATE $\ell_i=H(q_i,p_i)$ \% $H$ denotes cross entropy
\STATE $S=S\cup\{\log\ell_i\}$
\ENDFOR
\STATE Estimate GMM model on $S$  \hfill // Estimate GMM
\STATE Randomly initialize $f_\mathcal{D'}$
\FOR {epoch=1,2,$\cdots$,N}
\FOR {j=1,2,$\cdots$,k}
\STATE Sample a mini-batch data $X_j$ from $\mathcal{D'}$
\STATE Compute loss set $S_j$ on $X_j$ with $f_\mathcal{D}$
\STATE $\mathcal{W}_j = GMM(S_j)$
\STATE $A_j = \mathcal{W}_j*\alpha$
\STATE Compute loss $\mathcal{L}_{soft}$ on $X_j$ with $A_j$ as in \cref{eq:8}
\STATE Update $f_\mathcal{D'}$ with $\mathcal{L}_{soft}$
\ENDFOR
\ENDFOR  \hfill //ReSmooth Training
\RETURN $f_\mathcal{D'}$
\end{algorithmic}
\end{algorithm}

In ReSmooth stage, we adopt the original objective function for ID samples and slightly adjust the objective function for DAOOD samples. Specifically, we use label smoothing \cite{szegedy2016rethinking} strategy for DAOOD samples. In this way, we don’t change the marginal distribution $p(x)$ while changing the joint distribution $p(x,y)$. It has two properties: (1) keeping the diversity of DA and (2) not changing the separability in training.

Formally, we refer to $\boldsymbol{q}$ as the ground-truth one-hot label defined in the $\mathcal D'$ and $\boldsymbol{p}$ as the model prediction. Given the smoothness parameter $\alpha$ and the uniform distribution $\boldsymbol{u}$ over labels (satisfying $u_k=1/K$, where $K$ is the number of classes), the target label $\boldsymbol{q'}$ meets: $\boldsymbol{q}' = (1-\alpha) \boldsymbol{q} + \alpha \boldsymbol{u}$. So the cross-entropy loss with label smoothing (LS) is as follows:
\begin{align}
H(\boldsymbol{q'},\boldsymbol{p}) = (1-\alpha)H(\boldsymbol{q},\boldsymbol{p}) + \alpha H(\boldsymbol{u},\boldsymbol{p}),
\end{align}
where $H$ denotes cross-entropy.

With posterior probabilities $\{w_i\}$ derived from the detection stage, we can get the smooth parameter for each sample as $\alpha_i=\alpha(1-w_i)$. The soft split will give the sample not likely from original data distribution a higher $\alpha_i$ and the sample in original data distribution a lower $\alpha_i$. Our final loss for training on $\mathcal{D'}$ by adopting sample-wise label smoothing strategy comes as follows:

\begin{equation}
\label{eq:8}
\mathcal{L}_{div} = \frac{1}{|\mathcal D'|}\sum_{x_i\in \mathcal D'}(1-\alpha_i)H(\boldsymbol{q_i},\boldsymbol{p_i})+\alpha_i H(\boldsymbol{u},\boldsymbol{p_i}).
\end{equation}

The above loss is adopted for training with diverse DA. As for NDA, due to its OOD property at the beginning of design, we can ignore the detection procedure and simply utilize them in ReSmooth framework. To be specific, we assume all the augmented samples are OOD and use a constant hyperparameter $\alpha$ for all augmented samples for efficiency as follows:

\begin{align}
\label{eq:7}
\mathcal{L}_{neg} = &\frac{1}{|\mathcal D_{ID}'|}\sum_{x_i\in \mathcal D_{ID}'}H(\boldsymbol{q_i},\boldsymbol{p_i}) + \nonumber\\
&\frac{1}{|\mathcal D_{OOD}'|}\sum_{x_i\in \mathcal D_{OOD}'}(1-\alpha)H(\boldsymbol{q_i},\boldsymbol{p_i})+\alpha H(\boldsymbol{u},\boldsymbol{p_i}).
\end{align}

The overall algorithm can be found in \Cref{alg:resmooth}.

\section{Experiments}
\label{sec:Experiments}

To experimentally evaluate our proposed ReSmooth strategy, we first test it on several public vision datasets. Next, we validate the effectiveness of some designs in our framework by ablation study. Lastly, we conduct additional experiments to further address several potential concerns about DAOOD and DAID samples.

\subsection{Experimental Settings}
\begin{table*}[t]
\caption{Comparison between our approach and other baselines. All numbers in the table are top-1 accuracy (mean$\pm$std\%). {\bf RS+RA} indicates our ReSmooth strategy with RandAugment. Results in AA and KDforAA columns are from corresponding papers.}
\centering
\begin{tabular}{ccccccccccc}
\hline
Dataset                    & Model           & SA    & AA    & KDforAA  & RA    & LSR+RA & KDforRA & RS+RA \\ \hline
\multirow{2}{*}{CIFAR-10}  & Resnet18        & 95.41\tiny{$\pm$0.16} & -     & -        & 96.42\tiny{$\pm$0.12} & 96.38\tiny{$\pm$0.22}  & 96.59\tiny{$\pm$0.13}   & {\bf 96.65\tiny{$\pm$0.07}} \\
                           & WideResnet28-10 & 96.28\tiny{$\pm$0.05} & 97.40 & 97.60    & 97.61\tiny{$\pm$0.17} & 97.66\tiny{$\pm$0.07} 
& 97.45\tiny{$\pm$0.04}   & {\bf 97.84\tiny{$\pm$0.05}} \\
\multirow{2}{*}{CIFAR-100} & Resnet18        & 78.01\tiny{$\pm$0.34} & -     & -        & 78.84\tiny{$\pm$0.16} & 80.16\tiny{$\pm$0.23}  & 79.77\tiny{$\pm$0.10}   & {\bf 80.98\tiny{$\pm$0.18}} \\
                           & WideResnet28-10 & 80.96\tiny{$\pm$0.17} & 82.90 & 83.80    & 83.04\tiny{$\pm$0.16} & 83.87\tiny{$\pm$0.28} 
& 83.37\tiny{$\pm$0.07}   & {\bf 84.84\tiny{$\pm$0.16}} \\
\multirow{2}{*}{SVHN}      & Resnet18        & 96.58\tiny{$\pm$0.12} & -     & -        & 97.69\tiny{$\pm$0.05} & 97.74\tiny{$\pm$0.02}  & -       & {\bf 97.76\tiny{$\pm$0.04}} \\
                           & WideResnet28-10 & 97.09\tiny{$\pm$0.07} & 98.10 & -        & 98.10\tiny{$\pm$0.03} & 98.20\tiny{$\pm$0.05} 
& -       & {\bf 98.21\tiny{$\pm$0.03}} \\
ImageNet                   & Resnet50        & 76.84 & 77.60 & -        & -     & 77.88  & -       & {\bf 78.27} \\ \hline
\end{tabular}
\label{tab:performance}
\end{table*}

\begin{table*}[t]
\caption{Negative DA results. 'J' indicates jigsaw transformation and ‘R' indicates rotation.}
\centering
\begin{tabular}{ccccccccccc}
\hline
Dataset                    & Model  & SA    & J & J+LSR & J+KD  & J+RS  & R & R+LSR & R+KD  & R+RS      \\ \hline
\multirow{2}{*}{CIFAR-10}  & Res18      & 95.41\tiny{$\pm$0.16} & 95.68\tiny{$\pm$0.17} & 95.74\tiny{$\pm$0.12} & 95.59\tiny{$\pm$0.11} & \textbf{96.02\tiny{$\pm$0.03}} & 95.30\tiny{$\pm$0.11}  & 95.27\tiny{$\pm$0.11} & 95.22\tiny{$\pm$0.10} & \textbf{95.90\tiny{$\pm$0.23}} \\
                           & Wide28-10  & 96.28\tiny{$\pm$0.05} & 96.68\tiny{$\pm$0.15} & 96.73\tiny{$\pm$0.10} & 96.49\tiny{$\pm$0.05} & \textbf{96.95\tiny{$\pm$0.03}}   & 96.58\tiny{$\pm$0.14}   & 96.47\tiny{$\pm$0.10} & 96.27\tiny{$\pm$0.11}   & \textbf{96.94\tiny{$\pm$0.09}} \\
\multirow{2}{*}{CIFAR-100} & Res18      & 78.01\tiny{$\pm$0.34} & 78.00\tiny{$\pm$0.22} & 78.95\tiny{$\pm$0.15} & 78.67\tiny{$\pm$0.22} & \textbf{79.56\tiny{$\pm$0.30}} & 76.67\tiny{$\pm$0.33}  & 77.23\tiny{$\pm$0.02} & 77.97\tiny{$\pm$0.12} & \textbf{78.23\tiny{$\pm$0.32}} \\
                           & Wide28-10  & 80.96\tiny{$\pm$0.17} & 81.28\tiny{$\pm$0.11} & 82.25\tiny{$\pm$0.24} & 82.30\tiny{$\pm$0.17} & \textbf{82.71\tiny{$\pm$0.19}}   & 81.64\tiny{$\pm$0.17}   & 81.85\tiny{$\pm$0.29} & 81.86\tiny{$\pm$0.13}   & \textbf{82.27\tiny{$\pm$0.28}} \\ \hline
\end{tabular}
\label{tab:NDA}
\end{table*}

\begin{table*}[t]
\caption{Ablation Studies. We ablate our model with different datasets and network architectures.}
\centering
\begin{tabular}{c|c|ccc|ccc|cc}
\hline
Dataset & Model & Uni\_given & Uni\_avg & Uni\_optimal  & R.Sampling  & R.Split  & Reverse  & RS\_norm          & RS\_log \\ \hline
\multirow{2}{*}{CIFAR-10}& Res18    & 96.24\tiny{$\pm$0.11}  & 96.35\tiny{$\pm$0.14}  & 96.38\tiny{$\pm$0.22}  & 96.35\tiny{$\pm$0.10}  & 96.47\tiny{$\pm$0.06}  & 96.34\tiny{$\pm$0.07}    & \textbf{96.65\tiny{$\pm$0.10}}    & \textbf{96.65\tiny{$\pm$0.07}}\\ 
& Wide28-10                         & 97.31\tiny{$\pm$0.09}  & 97.61\tiny{$\pm$0.04}  & 97.66\tiny{$\pm$0.07}  & 97.60\tiny{$\pm$0.06}  & 97.71\tiny{$\pm$0.07}  & 97.56\tiny{$\pm$0.07}    & 97.73\tiny{$\pm$0.04}            & \textbf{97.84\tiny{$\pm$0.05}}\\
\multirow{2}{*}{CIFAR-100} & Res18  & 79.71\tiny{$\pm$0.41}  & 80.02\tiny{$\pm$0.46}  & 80.16\tiny{$\pm$0.23}  & 79.82\tiny{$\pm$0.10}  & 80.02\tiny{$\pm$0.11}  & 79.89\tiny{$\pm$0.15}    & 80.77\tiny{$\pm$0.29}             & \textbf{80.98\tiny{$\pm$0.18}}\\ 
& Wide28-10                         & 83.12\tiny{$\pm$0.14}  & 83.53\tiny{$\pm$0.18}  & 83.87\tiny{$\pm$0.28}  & 83.77\tiny{$\pm$0.11}  & 83.71\tiny{$\pm$0.18}  & 84.10\tiny{$\pm$0.09}    & 84.41\tiny{$\pm$0.15}             & \textbf{84.84\tiny{$\pm$0.16}}\\
\multirow{2}{*}{SVHN} & Res18       & 97.69\tiny{$\pm$0.05}  & 97.72\tiny{$\pm$0.01}  & 97.74\tiny{$\pm$0.02}  & 97.74\tiny{$\pm$0.07}  & 97.70\tiny{$\pm$0.07}  & 97.71\tiny{$\pm$0.01}    & \textbf{97.77\tiny{$\pm$0.01}}    & 97.76\tiny{$\pm$0.04} \\ 
& Wide28-10                         & 98.14\tiny{$\pm$0.04}  & 98.17\tiny{$\pm$0.03}  & 98.20\tiny{$\pm$0.05}  & 98.20\tiny{$\pm$0.02}  & 98.13\tiny{$\pm$0.03}  & 98.12\tiny{$\pm$0.03}    & 98.18\tiny{$\pm$0.01}             & \textbf{98.21\tiny{$\pm$0.03}}\\ \hline
\end{tabular}
\label{tab:ablation}
\end{table*}

\begin{table}[t]
\caption{A fair comparison experiment to illustrate the effectiveness of DAOOD samples. We follow \Cref{alg:fair} to train the model and the DAOOD/DAID test set is collected according to \Cref{alg:test}}
\centering
\begin{tabular}{cccc}
\hline
Data   & Test           & DAID Test         & DAOOD Test        \\ \hline
DAOOD  & 78.00          & 95.91             & \textbf{78.09}    \\
DAID   & 79.31          & \textbf{98.85}    & 48.14             \\
mix    & 79.73          & 96.49             & 77.40             \\ \hline
mix+RS & \textbf{80.09} & 97.35             & 77.31             \\ \hline
\end{tabular}
\label{tab:DAOOD}
% \vspace{0.2cm}
\end{table}

\begin{table}[t]
\caption{Experiments on more data augmentation strategies. 'X' below indicates the augmentation strategy without LSR or RS. Baseline indicates the training with standard augmentation.}
\centering
\begin{tabular}{cccc}
\hline
Method                  & X     & X+LSR & X+RS  \\ \hline
Baseline                & 78.01\tiny{$\pm$0.34} & -     & -     \\
AutoAugment             & 78.99\tiny{$\pm$0.14} & 80.14\tiny{$\pm$0.22} & \textbf{80.90\tiny{$\pm$0.16}} \\
Cutout ($8\times8$)     & 78.21\tiny{$\pm$0.21} & 78.92\tiny{$\pm$0.20} & \textbf{79.84\tiny{$\pm$0.18}} \\
Cutout ($16\times16$)   & 77.30\tiny{$\pm$0.26} & 78.76\tiny{$\pm$0.20} & \textbf{79.68\tiny{$\pm$0.24}} \\ 
TeachAugment            & 79.89\tiny{$\pm$0.33} & 80.92\tiny{$\pm$0.14} & \textbf{81.21\tiny{$\pm$0.19}} \\
\hline
\end{tabular}
\label{tab:others}
\end{table}

We evaluate our proposed method on four vision classification datasets: CIFAR-10~\cite{krizhevsky2009learning}, CIFAR-100~\cite{krizhevsky2009learning}, SVHN~\cite{netzer2011reading} and ImageNet datasets~\cite{deng2009imagenet}. CIFAR-10 dataset consists of 60K 32$\times$32 RGB images (6K images per class). CIFAR-100 dataset is similar to CIFAR-10, except that it has 100 classes and 600 images per class. SVHN is a dataset of house numbers, which has a core training set of 73K images and 531K additional training images. ImageNet is a large visual dataset consisting of natural images in high-resolution. We use the ImageNet-1000 subset of it, which has 1K classes and 1.3K training images and 50 validation images per class. In our experiments, we use the standard train split for the training but use the standard validation set in company with DAOOD and DAID validation set derived from it for testing in some experiments. The detailed description of DAOOD and DAID validation set can be found in \Cref{sec:dataset}. 

The backbone networks for training CIFAR-10, CIFAR-100 and SVHN are a small backbone ResNet-18~\cite{he2016deep} and a bigger and wider one WideResNet28-10~\cite{zagoruyko2016wide}. For ImageNet, we use ResNet-50~\cite{he2016deep} as the backbone. The implementation details are the same as \cite{cubuk2020randaugment}. Specifically, the ResNet-18 and WideResNet28-10 models were trained for 200 epochs (on CIFAR-10, CIFAR-100 and SVHN) with a learning rate of 0.1, batch size of 256, weight decay of 5e-4, and cosine learning rate decay.

The ResNet-50 models were trained for 180 epochs. The image size was 224 by 224, the weight decay was 2e-5 and the momentum optimizer with a momentum parameter of 0.9 was used. The learning rate was 0.1, which gets scaled by the batch size divided by 256. A global batch size of 256 was used, split across 2 workers.

In all experiments, we use an RTX 2080Ti GPU to train ResNet-18 and WideResNet28-10 and two RTX 2080Ti GPU to train the ResNet-50 model in parallel and amp. We report all the performance in the form of top-1 accuracy (mean$\pm$std\%) based on the average of three random runs except for experiments on ImageNet where we report results of one run with constant seed for all methods.

We pretrain $f_D$ with standard augmentation (detailed below) and train $f_D'$ with RA. For NDA, we consider CIFAR-10 and CIFAR-100 datasets with ResNet-18 and WideResNet28-10 models. We report the results for both the jigsaw transformation and the rotation transformation. We compare our work with several baselines to demonstrate the effectiveness of our proposed label smoothing strategy. The comparison baselines are briefly introduced as follows:\\
{\bf Standard Augmentation (SA):} a weak baseline. It includes random crop and random horizontal flip for CIFAR and SVHN while for ImageNet, it includes random resized crop, random horizontal flip and color jittering. "Unaugmented" data in this paper refers to the data with SA. \\
{\bf AutoAugment~\cite{cubuk2018autoaugment} (AA):} a strong baseline with AutoAugment.\\
{\bf RandAugment~\cite{cubuk2020randaugment} (RA):} a strong baseline with RandAugment.\\
{\bf RandAugment with Label Smooth~\cite{szegedy2016rethinking} (LSR+RA):} a strong baseline with RandAugment and adequate label smoothing strength.\\
{\bf KDforAA~\cite{wei2020circumventing}:} a state-of-the-art method for handling diverse DA by virtue of the model pretrained on un-/weak-augmented data.\\
{\bf KDforRA~\cite{wei2020circumventing}:} our reimplementation of \cite{wei2020circumventing} with RandAugment (using the same hyperparameters).

\subsection{DAOOD Dataset}
\label{sec:dataset}

\begin{algorithm}
\caption{Collect DAID and DAOOD dataset}
\label{alg:test}
\begin{algorithmic}[1]
\REQUIRE Pretrained $f_\mathcal{D}$, Train/test Set $\mathcal{D}$, $N$, $\mathcal{A}$, $GMM$
\ENSURE $\mathcal{D'_{ID}}$, $\mathcal{D'_{OOD}}$
\STATE $\mathcal{D'}_{ID} = \varnothing, \mathcal{D'}_{OOD} = \varnothing$
\WHILE {$|\mathcal{D'}_{ID}|$ or $|\mathcal{D'}_{OOD}| < N$}
\STATE Sample $(x_i, y_i)$ from $\mathcal{D}$  
\STATE Get $(x'_i, y'_i)$ by augmenting $(x_i, y_i)$ with $\mathcal{A}$
\IF {$y_i$ equals $f_\mathcal{D}(x_i)$}
\IF {$GMM(\ell(x'_i,y'_i)) < 0.5$ and $|\mathcal{D'}_{OOD}| < N$}
\STATE $\mathcal{D'_{OOD}}$ = $\mathcal{D'_{OOD}}\cup (x'_i,y'_i)$
\ENDIF
\IF {$GMM(\ell(x'_i,y'_i)) > 0.5$ and $|\mathcal{D'}_{ID}| < N$}
\STATE $\mathcal{D'_{ID}}$ = $\mathcal{D'_{ID}}\cup (x'_i,y'_i)$
\ENDIF
\ENDIF
\ENDWHILE
\RETURN $\mathcal{D'_{ID}}$, $\mathcal{D'_{OOD}}$
\end{algorithmic}
\end{algorithm}

Though the main purpose of this work is to improve the accuracy on the original test set when training with DA, we also collect the DAOOD dataset for potential further research like OOD generalization~\cite{djolonga2021robustness}. DAOOD samples are semantic-invariant under the meaning of recognition and should be generalized in robust model. Recall that DAOOD samples are samples that can't be recognized by the model $f_{\mathcal D}$ but can be recognized before DA. In accordance with it, we propose to select DAOOD samples from detected OOD samples by comparing the prediction result between original and augmented samples: if the original sample can be recognized but its augmented analogue can't, we view it as the DAOOD sample. We repeat this process until getting enough samples. The overall algorithm can be seen in \Cref{alg:test}. To be noticed, we do not use this algorithm for online detecting DAOOD samples in \Cref{sec:du} mainly because of its higher computation and memory cost. We collect three datasets, named CIFAR-10-RA, CIFAR-100-RA and ImageNet-RA, from CIFAR-10~\cite{krizhevsky2009learning}, CIFAR-100~\cite{krizhevsky2009learning} and ImageNet~\cite{deng2009imagenet}, respectively. The algorithm can be used to collect from any dataset combined with any DA strategy. The DAOOD dataset also have train and test split collected from its counterpart in the referenced dataset. It is to be noticed that there is no guarantee that any sample in referenced dataset has a counterpart in DAOOD dataset. What is more, even if the \Cref{alg:test} can select DAOOD samples more accurately than \Cref{alg:resmooth}, the collected dataset is still noisy.

\subsection{Classification Performance}

\subsubsection{Diverse DA}
From \Cref{tab:performance}, we can see that our method consistently improves on the basic DA strategy RA w or w/o LSR. To be specific, in CIFAR-10 our method achieves approximately 0.2\% improvement in comparison with the second-best, which is not trivial considering the baseline is relatively high. For CIFAR-100, the absolute improvement is 1.8\% compared to RA baseline and 1.0\% compared to the second-best method. 

On the other hand, however, the improvement in SVHN and ImageNet is relatively limited and the reasons behind them are slightly different. For SVHN, digits recognition is a comparatively easy task and mainly solved by the recognition of shape. In that sense, DAOOD samples are rarely detected in SVHN and most of the detected samples are semantics missing samples. A similar phenomenon can also be observed in ImageNet, but there still exists a large proportion of DAOOD samples. We will discuss the detected OOD samples later in detail in \Cref{sec:ood}.

\subsubsection{Negative DA}
We also augment the training data with NDA and train the model with different baselines. From \Cref{tab:NDA}, we can draw that using NDA in the usual way (w/ or w/o LSR) may not improve the generalization, especially for a small model. A smaller model has a poorer representative ability and may be hard to represent the data from two different distributions. With ReSmooth strategy, however, the model tends to learn better in the original data distribution while benefiting from the NDA, and the performance is consistently boosted. In addition, the KD-based method is not proposed to handle DAOOD samples and its performance is inconsistent and not as good as ReSmooth in the NDA setting.

\subsubsection{More Data Augmentation Strategies}

We carry out some quick experiments on more data augmentation strategies in this section to verify the consistent improvement of our proposed ReSmooth framework. We consider two well-known DA methods AA\cite{cubuk2018autoaugment} and Cutout\cite{devries2017improved}, and a recently proposed generation-based DA method TeachAugment\cite{suzuki2022teachaugment}. Cutout augmentation randomly cut a patch out from the original image and the residual parts can be viewed as semantic-unchanging. We consider two cut sizes: $8\times8$ and $16\times16$. TeachAugment trained an augmentation network and a recognition network in an adversarial manner, where the training of augmentation network is constrained by a teacher model. We have experiments on CIFAR-100 with Resnet18 and the results can be found in \Cref{tab:others}. We can see that ReSmooth can be easily extended to more augmentations and improve on them.

\begin{figure*}[t]
  \centering
  \includegraphics[width=1.0\linewidth]{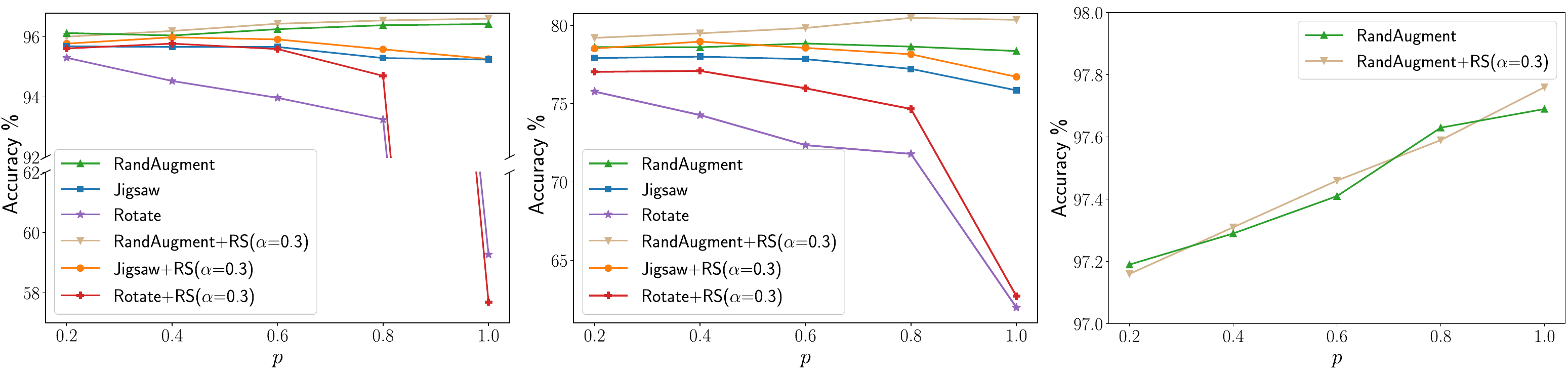}
  \vspace{-0.6cm}
  \caption{The influence of the augmentation probability $p$ on different datasets with Resnet18. Left: The influence of $p$ on CIFAR-10 with NDA and diverse DA. Middle: The influence of $p$ on CIFAR-100 with NDA and diverse DA. Right: The influence of $p$ on SVHN with diverse DA.}
  \label{fig:Psearch}
\end{figure*}
\begin{figure*}[t]
  \centering
  \includegraphics[width=1.0\linewidth]{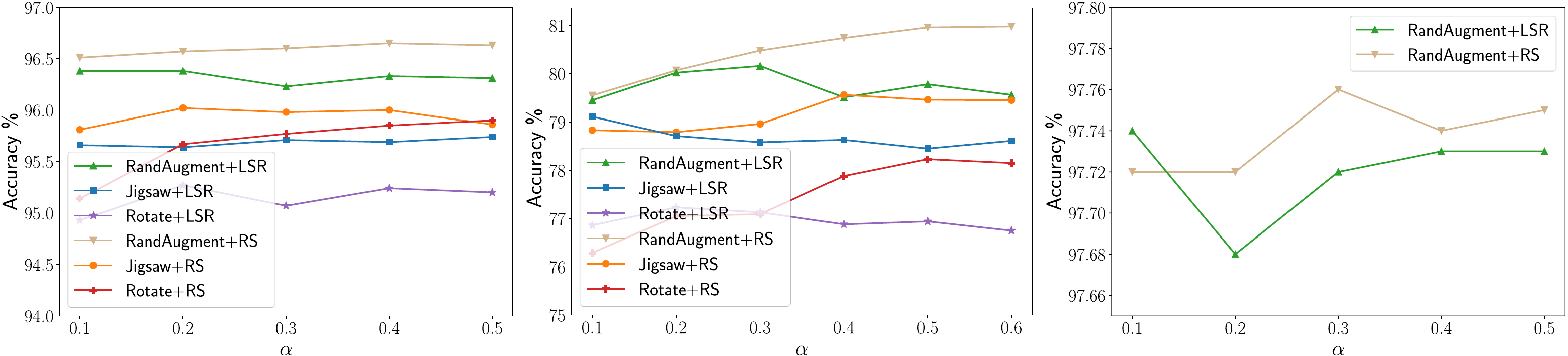}
  \vspace{-0.6cm}
  \caption{The influence of the smoothing parameter $\alpha$ on different datasets with Resnet18. Left: The influence of $\alpha$ on CIFAR-10 with NDA and diverse DA. Middle: The influence of $\alpha$ on CIFAR-100 with NDA and diverse DA. Right: The influence of $\alpha$ on SVHN with diverse DA.}
  \label{fig:Asearch}
\end{figure*}

\subsection{Ablation Study}
\label{Sec:ablation}

The only difference between our method and LSR+RA baseline in training is the label smoothing parameters. In LSR+RA baseline, one label smoothing parameter $\alpha$ is used for all training samples, whereas in our method, different samples with different transformations are given different label smoothing parameters based on the posterior probabilities estimated in the OOD detection stage: $\alpha_i=\alpha(1-w_i)$. To validate the effectiveness of ReSmooth framework, we ablate the choice of the label smoothing parameters. Specifically, we compare our detection-based label smoothing parameters with the following methods: (1) Uni\_given. Unified smoothness where the parameter is the same as the (max) smooth parameter of ReSmooth framework, (2) Uni\_avg. Unified smoothness where the parameter equals the average of the smooth parameters in ReSmooth framework, (3) Uni\_optimal. Unified smoothness where the parameter is optimal in a grid search step, (4) Random Sampling. Give every sample a random smooth parameter sampling from a uniform distribution $\mathcal U(0,\alpha)$, (5) Random Split. Shuffle the smooth parameters derived from the OOD detection step, (6) Reverse. Give every sample a reversed smooth parameter: $\alpha_i=\alpha w_i$.

On the other hand, we ablate the loss distribution estimation by replacing the log loss distribution with the normalized loss distribution. In this situation, the loss distribution is unimodal and the estimated GMM is arguably unimodal (because one Gaussian component is centered while the other Gaussian component is not and comparably negligible). Based on the posterior probability distribution of the centered Gaussian component, we follow ReSmooth framework to train the new network. The results can be found in "RS\_norm" column and "RS\_log" column.

Results in \Cref{tab:ablation} suggest that though smoothed labels can always benefit the performance, our methods are the best and improve on them with a large margin. Noticeably, sample-wise strategies are not of consistent performance across different datasets except for ours. It suggests that sample-wise smooth heavily relies on the smooth parameters and our parameters derived from OOD detection are helpful and appropriate for all datasets. What is more, the performance of ReSmooth combined with the normalized loss distribution is similar but inferior to the best one. It is because that the estimated Gaussian distribution is centered and thus the sample deviated from the center tend to be assigned a bigger smooth parameter. It has two disadvantages: (1) more ID-like samples are viewed as OOD samples and they are less distinguishable; (2) the training is attenuated because of the bigger label smoothing parameters. As a result, the performance of RS\_norm is sub-optimal and the gap becomes even larger for bigger model.

\subsection{Parameter Analysis}

Given a specific data augmentation strategy, the only hyperparameter to tune in ReSmooth framework is the smooth parameter $\alpha$. What is more, we observe that sometimes our ReSmooth framework perform better with larger augmentation probability $p$ compared to baselines. As a result, we perform two separate search of the optimal hyperparameters: $\alpha$ and $p$. At the beginning, we fixed the smooth parameter $\alpha$ and search $p$, and then we fix $p$ and search $\alpha$. Because the optimal smooth parameter and the optimal augmentation probability are influenced by both the dataset\cite{muller2019does} and the augmentation strategy, we perform the search process independently per augmentation strategy per dataset as shown in \Cref{fig:Psearch}. Results suggest that for diverse DA, bigger $p$ usually corresponds to better performance while for NDA, bigger $p$ causes worse performance in standard training. Compared to it, ReSmooth framework can always ameliorate the performance and benefit from the larger augmentation probability.

Once $p$ is determined, we turn to find the optimal $\alpha$. In this section, we provide the results of both ReSmooth and the standard training scheme with LSR to illustrate the different appearances of them. $p$ for LSR is set as same as the optimal $p$ for standard training, because we find it performs empirically better under this setting. Results in \Cref{fig:Asearch} show that ReSmooth framework is superior to the standard training scheme with LSR in a large range of $\alpha$. What is more, ReSmooth usually benefit from larger $\alpha$ compared to the LSR baseline. The above results and observations suggest that setting $\alpha\in[0.3, 0.5]$ is suitable for most of the cases across different datasets and augmentations. For ImageNet results, we only search $\alpha$ with the same augmentation strategy as \cite{cubuk2020randaugment} and we tried 0.3, 0.4, 0.5. The accuracies are 77.98, 78.09 and 78.27 respectively. For WideResnet28-10 results, we experimented with the searched $p$ of Resnet18 and $\alpha$s (if exist) that result in similar performance with Resnet18. Almost all the optimal hyperparameters of Resnet18 are also optimal for WideResnet28-10 except that the optimal $\alpha$ of WideResnet28-10 for jigsaw on CIFAR-10 is 0.3 while $\alpha$ is 0.2 for Resnet18.

\subsection{Potential Concerns}
\begin{figure*}
    \centering
    \subfigure[ImageNet]{
        \begin{minipage}[t]{\linewidth}
            \centering
            \includegraphics[width=0.9\textwidth]{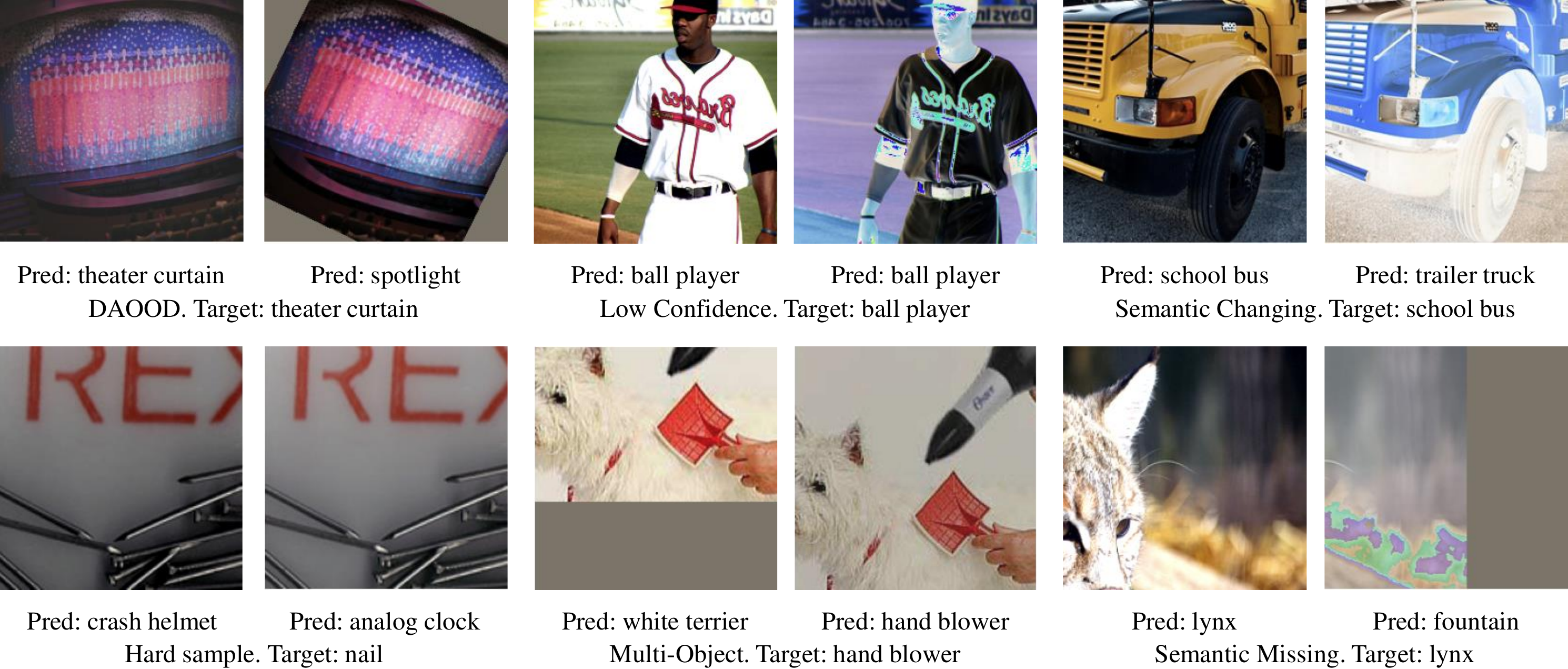}
        \end{minipage}
    }
    \hfill
    \subfigure[CIFAR-100]{
        \begin{minipage}[t]{\linewidth}
            \centering
            \includegraphics[width=0.9\textwidth]{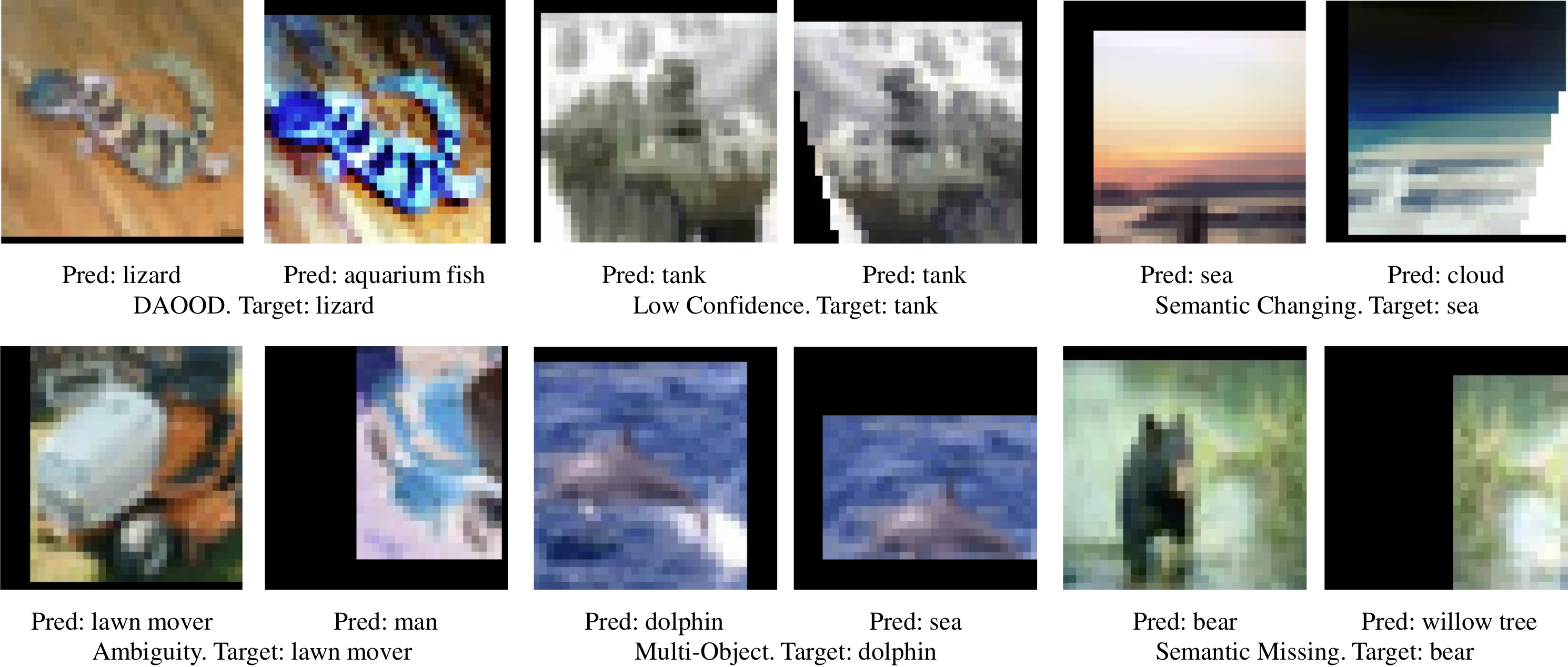}
        \end{minipage}
    }
    \hfill
    \subfigure[SVHN]{
        \begin{minipage}[t]{\linewidth}
            \centering
            \includegraphics[width=0.9\textwidth]{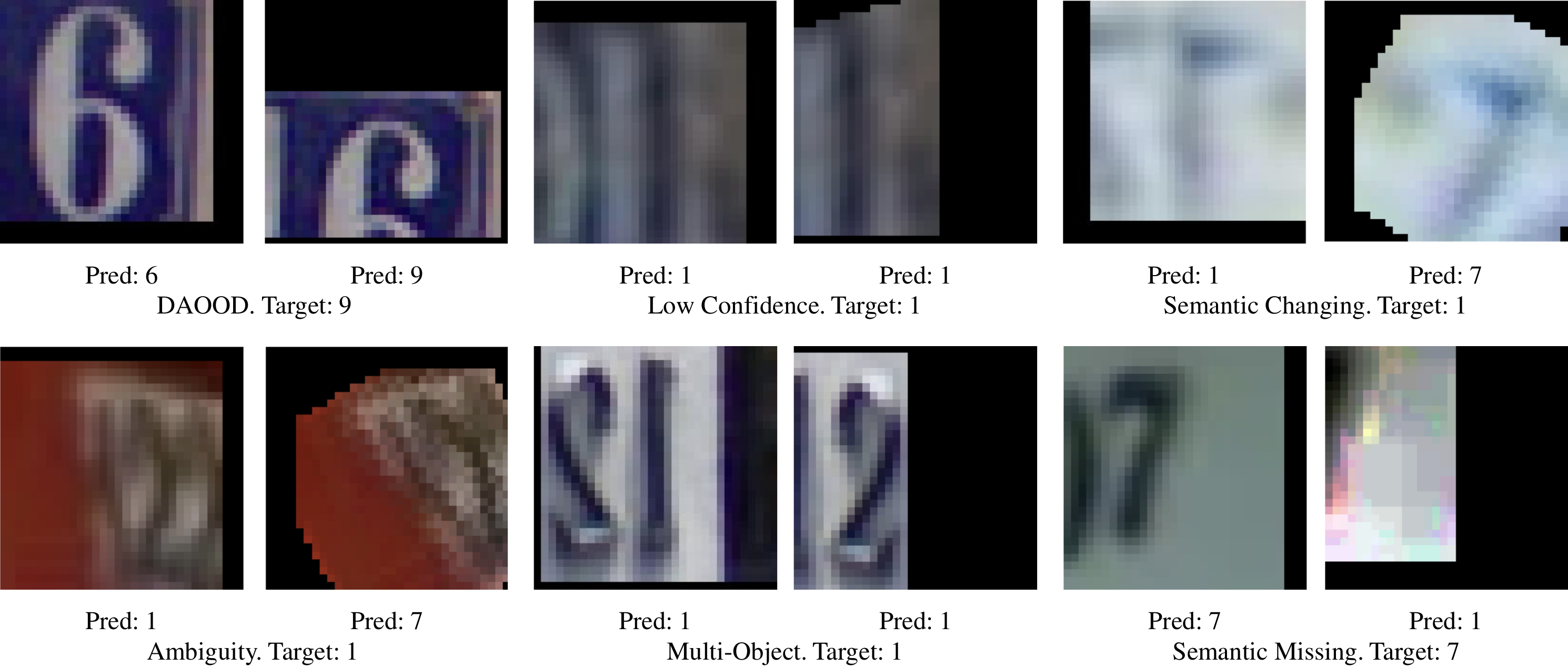}
        \end{minipage}
    }
    \caption{Different types of detected OOD samples on ImageNet, CIFAR-100 and SVHN training set. For each pair, we show the image before RandAugment (left) and the image after RandAugment (right). Under each image is their corresponding label predicted by the pretrained model.}
    \label{fig:OODs}
\end{figure*}

In this section, we will figure out some potential concerns about DAID and DAOOD samples empirically.

\begin{figure*}[t]
  \centering
  \includegraphics[width=1.0\linewidth]{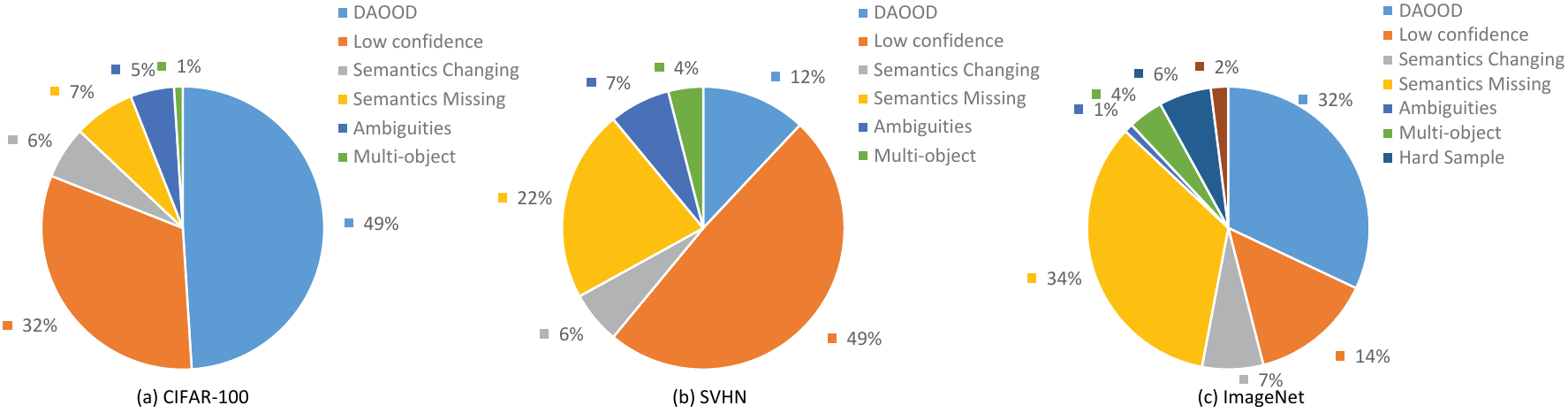}
  \caption{Ingredients of detected OOD samples on each dataset considered. The number of DAOOD samples in the augmented data is distributed differently on different datasets, which leads to different performances of ReSmooth on these datasets.}
  \label{fig:proportion}
\end{figure*}

\subsubsection{Are DAOOD Samples Needed for Training?}
\label{sec:need}

On one hand, DAOOD samples are samples out of the original data distribution, and learning from it may confuse the model, change the data manifold, and consequently diverts the representation learning toward the DAOOD samples. On the other hand, either from the perspective of DA diversity or based on the intuition that DAOOD samples are informative and semantic-preserving, DAOOD samples may help train a model of better generalization. To verify the effectiveness of DAOOD during training, we propose to fairly compare training with DAID, training with DAOOD, and training with the mixture of them as usual. The main principle we are concerned with is to use the same number of samples of DAID, DAOOD, and mix to train the models respectively as shown in \Cref{alg:fair}. The results can be found in \Cref{tab:DAOOD}. 

From it, we can find that training with DAID samples is the easiest and the trained model performs well in DAID test set and original test set. It is expected because DAID samples are supposed to lie in the same distribution as the original data. On contrary, the model trained on DAOOD performs worse (but not too bad) in both the original test set and DAID test set. The model trained on the mixture, though not performing best in DAID or DAOOD test set, achieves best test set accuracy. On top of that, our method ameliorates the training with the mixture and has better performance in both the DAID and original test set while incurring a negligible decrease in DAOOD test set. Therefore, we conclude that DAOOD samples help the model generalization and our method can make better use of these DAOOD samples.

\subsubsection{What Samples are Detected?}
\label{sec:ood}

\begin{algorithm}[t]
\caption{Fair Comparison Training}
\label{alg:fair}
\begin{algorithmic}[1]
\REQUIRE $f_\mathcal{D}$, $GMM$, $\mathcal{D}$, $FLAG$, maximal samples $n$
\ENSURE $f_\mathcal{D'}$
\STATE Randomly initialize $f_\mathcal{D'}$. 
\FOR {epoch=1,2,$\cdots$,N}
\STATE Sample a mini-batch data $X_j$ from $\mathcal{D'}$
\STATE Split $X_j$ into $X_{ID}$ and $X_{OOD}$ given $f_\mathcal{D}$ and $GMM$
\STATE $m = \min(n,|X_{ID}|,|X_{OOD}|)$
\IF {$FLAG$ is "DAOOD"}
\STATE Randomly get $m$ samples ($S_j$) from $X_{OOD}$
\ELSIF {$FLAG$ is "DAID"}
\STATE Randomly get $m$ samples ($S_j$) from $X_{ID}$
\ELSIF {$FLAG$ is "mixture"}
\STATE Randomly get $m$ samples ($S_j$) from $X_j$
\ENDIF
\STATE Compute loss $\mathcal{L}$ on $S_j$ with $f_\mathcal{D'}$ as in \cref{eq:1}
\STATE Update $f_\mathcal{D'}$ with $\mathcal{L}$
\ENDFOR
\RETURN $f_\mathcal{D'}$
\end{algorithmic}
\end{algorithm}

To verify the effectiveness of the OOD detection module, we try to figure out what kinds of samples are detected (distinguished by a threshold $\tau$) and their corresponding proportions in this section. We show some examples of detected samples along with their predicted labels by the pretrained model and the samples before DA to better identify different types of them. These examples can be found in \Cref{fig:OODs}. Specifically, we have divided samples into seven categories: (1) DAOOD sample, as described in \Cref{sec:daood}, (2) low confidence sample, which is semantic-unchanging and is rightly predicted after augmentation but with a low confidence, (3) hard sample, which is semantic-unchanging but is wrongly predicted after augmentation, (4) semantics changing sample, where the object to be recognized remains in the image but semantic label changes after augmentation (e.g., a sea changes to a cloud in \Cref{fig:OODs}), (5) semantics missing sample, where the object misses after augmentation, (6) ambiguity, which is not recognizable for us, (7) multi-object sample, where multiple objects exist and multiple labels are reasonable. In addition, we estimate their proportions by random sampling one hundred samples. Results are shown in \Cref{fig:proportion}. We find that the situation differs between CIFAR-100 dataset and ImageNet. In CIFAR-100, most of OOD samples detected are DAOOD samples while most of the remaining are predicted correctly with low confidence. In ImageNet, the sources of data are heterogeneous, including DAOOD samples, semantics changing samples, semantics missing samples, multi-object scene, hard samples, and ambiguity samples.

Intuition suggests that samples with low confidence, including both the low confidence samples (with right prediction) and hard samples (with wrong prediction), can be viewed as the DAOOD samples or the OOD samples in the original datasets (if samples before DA are also with low confidences) to some extent. In this sense, we argue that our method is suitable to handle both the DAOOD samples and samples with low confidence, but can't do well in other situations like missing semantics (e.g., noisy label or not cropping informative region) or multi-object scene recognition. We think this is the main reason why the improvement is relatively limited in SVHN and ImageNet datasets. On the other hand, the proportions may uncover the underlying reason why distillation-based methods can improve performance like \cite{wei2020circumventing}. Distillation-based methods can ameliorate training when the given label is noisy or insufficient. But on the contrary, distillation-based methods are not good at handling DAOOD samples and samples with low confidence for the teacher model will provide wrong supervision on these samples. As a result, it is reasonable to find that our method perform better than \cite{wei2020circumventing} when DAOOD samples, low confidence samples and hard samples are of a large proportion. In other word, though the claimed phenomenon in \cite{wei2020circumventing} can be observed by us, it is not the most case. Naturally, a fine-grained split of detected OOD is preferred to better facilitate the learning problem with diverse DA. Combining the predictions before and after augmentation may be helpful because DAOOD samples are predicted correctly before augmentation and wrongly after augmentation. But this method is time- and memory-consuming, so we leave the fine-grained split problem into future research.

\section{Conclusion}
\label{sec:Conclusions}
In this paper, we propose ReSmooth framework to ameliorate the training with diverse DA. The key observation behind it is that a big part of the augmented samples are out of the original data distribution, denoted as DAOOD samples. We empirically show that DAOOD samples are crucial for training a model of better generalization and our ReSmooth framework can better utilize DAOOD samples and further improve on regular training strategy. In addition, we also include the learning with NDA as part of work that only involves OOD utilizing. Classification results show that in most of the cases, our proposed framework can benefit the learning of both diverse DA and negative DA.

%{\appendices
%\section*{Proof of the First Zonklar Equation}
%Appendix one text goes here.
% You can choose not to have a title for an appendix if you want by leaving the argument blank
%\section*{Proof of the Second Zonklar Equation}
%Appendix two text goes here.}
 
 % argument is your BibTeX string definitions and bibliography database(s)

\bibliographystyle{IEEEtran}
\bibliography{TNNLS-2022-P-21693}

%\bibliography{IEEEabrv,../bib/paper}
%

\newpage

\vspace{11pt}

\vfill

\end{document}